\documentclass[preprint,12pt,authoryear]{elsarticle}

\usepackage{amssymb}
\usepackage{hyperref}
\usepackage{lineno}

\usepackage{amsmath} 

\usepackage{array}
\usepackage{todonotes}
\usepackage{natbib}
\usepackage{booktabs}

\usepackage{enumitem}

\usepackage{geometry}
\journal{Elsevier}

\begin{document}

\begin{frontmatter}



\title{Scalable and Reliable Multi-agent Reinforcement Learning for Traffic Assignment}


\author[a]{Leizhen Wang}
\author[a]{Peibo Duan\corref{cor1}}  
\author[b]{Cheng Lyu}
\author[c]{Zewen Wang}
\author[d]{Zhiqiang He}
\author[e]{Nan Zheng}

\author[f]{Zhenliang Ma\corref{cor1}}  

\cortext[cor1]{Corresponding author.}

\affiliation[a]{organization={Department of Data Science and Artificial Intelligence, Monash University},
            city={Melbourne},
            country={Australia}}
\affiliation[b]{organization={Chair of Transportation Systems Engineering, Technical University of Munich},
            city={Munich},
            country={Germany}}

\affiliation[c]{organization={School of Transportation, Southeast University},
            city={Nanjing},
            country={People's Republic of China}}

\affiliation[d]{organization={The Graduate School of Informatics and Engineering, The University of Electro-Communications},
            city={Tokyo},
            country={Japan}}
\affiliation[e]{organization={Department of Civil Engineering, Monash University},
            city={Melbourne},
            country={Australia}}

\affiliation[f]{organization={Department of Civil and Architectural Engineering, KTH Royal Institute of Technology},
            city={Stockholm},
            country={Sweden}}

\begin{abstract}

The evolution of metropolitan cities and the increase in travel demands impose stringent requirements on traffic assignment methods. Multi-agent reinforcement learning (MARL) approaches outperform traditional methods in modeling adaptive routing behavior without requiring explicit system dynamics, which is beneficial for real-world deployment. However, MARL frameworks face challenges in scalability and reliability when managing extensive networks with substantial travel demand, which limiting their practical applicability in solving large-scale traffic assignment problems. To address these challenges, this study introduces MARL-OD-DA, a new MARL framework for the traffic assignment problem, which redefines agents as origin-destination (OD) pair routers rather than individual travelers, significantly enhancing scalability. Additionally, a Dirichlet-based action space with action pruning and a reward function based on the local relative gap are designed to enhance solution reliability and improve convergence efficiency. Experiments demonstrate that the proposed MARL framework effectively handles medium-sized networks with extensive and varied city-level OD demand, surpassing existing MARL methods. When implemented in the SiouxFalls network, MARL-OD-DA achieves better assignment solutions in 10 steps, with a relative gap that is 94.99\% lower than that of conventional methods. 
\end{abstract}




\end{frontmatter}



\section{Introduction}
\label{sec:intro}

Traffic assignment is a pivotal element in traffic planning, concerned with the distribution of OD demand within a transport network to achieve user equilibrium (UE) or system optimal (SO) objectives, as informed by the system goals~\citep{boyles2020transportation,de2024modelling}~\footnote{UE focuses on minimizing travel costs for individuals, whereas SO aims to minimize the total network-wide costs.}. The rise of metropolitan areas and the increase in travel demands require advanced online decision making tools to provide route guidance to human drivers, connected and automated vehicles, and shared mobility on demand. These developments place growing pressure on the efficiency, scalability, and real-time performance of traffic assignment methods~\citep{luo2023alpharoute,zhou2022scalable,guo2021mixed,liu2020integrated,lujak2015route}.

Driven by recent advancements in Artificial Intelligence (AI) and data acquisition technologies, traditional traffic assignment methods have undergone substantial refinement. These approaches can be classified into two categories: supervised learning (SL)-based~\citep{rahman2023data,liu2024end} and MARL-based~\citep{shou2022multi,zhou2020reinforcement,ramos2018analysing}. SL-based methods predict UE flow patterns but overlook critical travel behaviors, such as route choice, and fail to ensure link-path flow relationships. In contrast, MARL-based approaches bridge the gap in SL-based models by viewing travelers as independent agents who learn from continuous interactions with the traffic environment.This enables them to respond to dynamic traffic conditions, operate under partial information, handle complex and non-convex traffic scenarios, and integrate individual behavior modeling with system-level optimization. Furthermore, to improve the performance of MARL-based methods, most studies focus on action regulation like en-route and route decisions.

Despite advancements in MARL frameworks have led to impressive performance through enhanced modeling capabilities, such as action regulation, they still encounter substantial hurdles, notably in terms of \textbf{scalability} and \textbf{reliability} (\textbf{S\&R}), which are mainly attributed to the intrinsic modeling approach. As existing MARL frameworks typically assign one agent per traveler, they often struggle to scale when the number of agents increases, given that real-world transport systems may involve huge numbers of travelers. For example, \citet{shou2022multi} proposed a mean field Q-learning approach to address scalability issues, yet their trials were limited to \textit{2100} travelers. Another study addressed the traffic assignment problem involving 360,600 travelers, but did not achieve the desired accuracy~\citep{bazzan2016multiagent}. Furthermore, the largest traffic networks in these methods comprise no more than \textit{69} nodes and \textit{166} links~ \citep{shou2022multi,zhou2020reinforcement,ramos2018analysing}, still representing small-sized networks insufficient for real-world application. In tandem with the scalability issue, flexibility has emerged as a critical concern. Due to the agent-per-traveler design, MARL frameworks rely on a fixed number of agents during training. However, in real-world scenarios, fluctuations in OD demand require adjustments in the agent configuration, necessitating model retraining or fine-tuning. This process incurs substantial computational overhead and introduces risks such as catastrophic forgetting, thereby limiting the practical deployment of MARL in dynamic traffic environments.

The reliability of MARL-based approaches is demonstrated through their ability to efficiently converge to a plausible solution applicable in real-world situations. However, many current investigations oversimplify real-world conditions by modeling travelers as making route choices solely based on location and time, neglecting the complexity of their decision-making processes, which compromises the quality of the solution. Furthermore, using average travel time to assess convergence to the UE works at cross purposes since it more likely aligns with the SO goal rather than minimizing individual costs. Braess' Paradox highlights that reaching UE can lead to increased average travel times compared to SO~\citep{zhuang2022braess}.

In order to tackle the previously mentioned issues of scalability and reliability, this paper re-examines the MARL framework applied to traffic assignment, encompassing the definition of agents and the tailored state-action-reward formulation. The main contributions are as follows:

\begin{itemize}
    \item \textbf{Model}: A novel MARL framework, \textbf{MARL-OD-DA}, is proposed, where each agent is redefined as an OD pair router responsible for routing policies across travelers sharing the same OD pair. This redesign significantly improves scalability by reducing agent complexity and enables adaptation to varying OD demand, while also enhancing learning reliability in large-scale networks.
    \item \textbf{Action and Reward Formulation}: A Dirichlet-based action space and a relative gap-based reward are introduced to better align learning with system efficiency. Additionally, an action pruning mechanism is proposed to enforce output sparsity and address the numerical instability of Dirichlet-based strategies, further improving reliability.
    \item \textbf{Validation}: The proposed method is validated on three transportation networks, ranging from small to medium-sized networks with variable OD demand. Results demonstrate significant improvements in scalability and reliability compared to existing methods.
\end{itemize}

The remainder of this paper is organized as follows. Section~\ref{sec:review} reviews conventional and learning-based traffic assignment methods. Section~\ref{sec: problem statement and analysis} outlines the problem and limitations of existing MARL frameworks. Section~\ref{sec:method} presents the proposed methodology. Section~\ref{sec:experiment} evaluates its performance on three transportation networks. Finally, Section~\ref{sec:conclusion} concludes the paper and outlines future research directions.

\section{Related Work}
\label{sec:review}

\subsection{Conventional Traffic Assignment Methods}

Traffic assignment problems are typically solved using mathematical programming methods. These methods can be categorized into three types: link-based~\citep{mounce2015convergence,fukushima1984modified,frank1956algorithm}, path-based~\citep{chen2020parallel,jayakrishnan1994faster}, and origin-based approaches~\citep{nie2010class,bar2002origin}, all of them follow an iterative workflow: first, calculate travel times, then identify the shortest paths, adjust route choices toward equilibrium, and repeat the process. In parallel, several studies have explored traffic assignment from the perspective of traveler behavior and day-to-day dynamics~\citep{siri2022topology,wang2024ai}. In recent years, significant efforts have been made to improve the computational efficiency of these methods~\citep{liu2024novel,zhang2023admm,chen2020parallel}. Partly inspired by the path-based method, this paper treats all OD-pairs as agents and performs optimization in parallel, offering a novel perspective to improve scalability and efficiency.

\subsection{Learning-based Traffic Assignment Methods}
\label{subsec: learning-based TA}

Recent studies have explored learning-based approaches for traffic assignment. For example, \citet{wang2024ai} proposed a large language model (LLM)-based agent to simulate human-like adaptive behavior in road networks, with the aim of achieving UE. However, their focus lies in behavior modeling rather than solving the assignment problem itself. In general, learning-based methods can be categorized into supervised learning (SL) and reinforcement learning (RL) approaches.

\textbf{SL-based methods.}
\citet{rahman2023data} introduced Graph Neural Networks (GNNs) for predicting UE assignment flows without making assumptions about travelers' routing behavior. Their experimental results demonstrated the ability to predict traffic flows with less than 2\% mean absolute error. \citet{liu2024end} further improved this approach by adding auxiliary loss functions, ensuring that the predicted results partially satisfy the constraint of link-path flow relationship. \citet{liu2024heterogeneous} extended this methodology to predict link flows without relying on sensor data. \citet{liu2023end} proposed a novel end-to-end learning framework that directly learns travelers' route choice preferences and the UE flow from observational data. While this method shows potential, it is currently limited to proof-of-concept experiments, and thus was not pursued further in this study. Although SL-based models show promising accuracy in flow prediction, they typically require large volumes of labeled data and often lack interpretability in representing travelers' decision processes, limiting their deployment in dynamic or data-scarce scenarios.

\textbf{RL-based Methods.} 
While SL-based methods primarily approximate UE flow patterns from observed data, RL-based methods instead aim to learn sequential decision-making policies through interaction with the environment. \citet{wang2025reinforcement} proposed a single-agent RL framework that recommends routes for all travelers to achieve SO, with the agent acting as a centralized decision-maker. In contrast, most existing studies adopt MARL, where each agent represents an individual traveler making decentralized routing decisions.

Based on the granularity of an agent's action, MARL-based methods can be categorized into two main types. The first is \emph{route-based} methods, in which an agent selects a complete route from a set of pre-defined candidates and strictly follows it from origin to destination~\citep{zhou2020reinforcement, ramos2018analysing, wang2025reinforcement}. These route sets are commonly generated using $k$-shortest path algorithms or extracted from historical trajectory data. The second category is \emph{en-route} methods, where an agent makes sequential decisions at each node by choosing one of the outbound links, continuing this process until the destination is reached or a predefined step limit is exceeded~\citep{shou2022multi, bazzan2016multiagent}. Compared to route-based approaches, en-route methods offer greater flexibility but often introduce higher complexity in policy learning and convergence.

Compared to conventional model-based traffic assignment techniques, MARL-based approaches offer several distinctive advantages. First, agents learn adaptive routing policies through repeated interactions, allowing them to respond to dynamic traffic conditions~\citep{zhou2020reinforcement, shou2022multi}. Second, RL-based methods are model-free and do not rely on explicit knowledge of system dynamics. This allows them to operate under partial or noisy observations and to solve complex, non-convex traffic scenarios that are often intractable for traditional optimization techniques. Finally, MARL naturally integrates individual behavior modeling with system-level optimization, eliminating the need for separately calibrated route choice models.

\section{Problem Statement and Analysis}
\label{sec: problem statement and analysis}

In traffic assignment, the urban transportation network is modeled as a directed graph \( G(V, E) \), where \( V \) represents the intersections (nodes) and \( E \) represents the road segments (edges). Each road segment \( e \in E \) is associated with a performance function \( c_e(x_e) \), which maps the flow \( x_e \) on the segment to its corresponding travel time. The travel demand between OD pairs is represented by a demand matrix \( D = \{d_{rs}\} \), where \( d_{rs} \) denotes the travel demand from origin node \( r \) to destination node \( s \).

The traffic assignment problem seeks to allocate this demand across the available routes in the network, typically with UE or SO objective. For example, the UE objective function is formulated as:
\begin{equation}
    \min_{x} \sum_{e \in E} \int_0^{x_e} c_e(v) \, dv
\end{equation}
These objectives are subject to the following constraints:
\begin{itemize}
    \item \textbf{Flow conservation.} The total flow distributed across all feasible paths between an origin \( r \) and a destination \( s \) must match the corresponding demand \( d_{rs} \):
    \begin{equation}
        \sum_{p \in P_{rs}} f_p = d_{rs}, \quad \forall (r, s) \in D,
    \end{equation}
    where \( P_{rs} \) represents the set of all valid paths between \( r \) and \( s \), and \( f_p \) denotes the flow assigned to path \( p \).
    \item \textbf{Non-negativity.} All path flows must remain non-negative to ensure feasibility:
    \begin{equation}
        f_p \geq 0, \quad \forall p \in P_{rs}, \, \forall (r, s) \in D
    \end{equation}
    \item \textbf{Link-path flow relationship.} The total flow on a road segment \( e \) is calculated as the sum of the flows on all paths that traverse \( e \):
    \begin{equation}
        x_e = \sum_{(r, s) \in D} \sum_{p \in P_{rs}: e \in p} f_p, \quad \forall e \in E
    \end{equation}
\end{itemize}

To this end, the MARL framework typically formulates the traffic assignment problem as a Decentralized Partially Observable Markov Decision Process (DEC-POMDP), which is well-suited for modeling complex multi-agent systems with local observations and decentralized objectives. Formally, a DEC-POMDP is defined by the tuple $\langle S, \mathcal{A}, \mathcal{O}, \mathcal{R}, P, n, \gamma \rangle$, where \( S \) represents the state. \( \mathcal{A} = A_1 \times A_2 \times \cdots \times A_n \) is the joint action space, with \( A_i \) denoting the action for agent \( i \). Similarly, \( \mathcal{O} = O_1 \times O_2 \times \cdots \times O_n \) is the joint observation space, where \( O_i \) represents the local observation of agent \( i \). \( \mathcal{R} = \{R_i\} \) is the set of rewards, where each \( R_i = R_i(S, \mathcal{A}) \) defines the reward for agent \( i \). The transition function \( P \) specifies the probability of transitioning from one state to another given the joint actions of all agents. The parameter \( n \) denotes the number of agents. Finally, \( \gamma \in [0, 1] \) is the discount factor, which balances the importance of immediate rewards and long-term objectives.\footnote{For a detailed review of DEC-POMDPs, refer to~\citep{oliehoek2016concise}.}

Existing MARL framework conceptualizes an agent as a traveler with a discrete action space, where each action corresponds to a chosen route. Thus, the count of agents equals the number of travelers, or equivalently, the size of OD demand. As discussed in Section~\ref{sec:intro}, S\&R problems arise within this framework. Conversely, this paper advocates for the recharacterization of the agent to function as an OD-pair router to address SAR challenges. Nonetheless, the subsequent issues must be resolved when establishing such a type of agent-focused MARL framework:

\begin{enumerate}[label=\roman*)]
    \item According to the re-defined agent in our MARL framework, a straightforward redefinition of the action space should reflect the potential number of travelers choosing a given route. Nevertheless, if this discrete action space formulation persists, two issues arise: firstly, the action space dimensions for an agent become inconsistent; secondly, the vast number of travelers opting for that route results in a massive action space size. Consequently, these conditions perpetuate the dilemma associated with the extensive scale of OD demand.

    \item The relative gap serves as a more accurate measure of convergence, as detailed in formula~\eqref{eq:global_RG} within Section~\ref{sec:reward}. However, its conventional computation relies on global traffic information, which is often inaccessible to individual travelers in real-world settings. Since agents in practical deployments typically receive only local feedback, it becomes crucial to devise a localized variant of the relative gap that can be computed from partial observations while still serving as a meaningful convergence indicator.
\end{enumerate}

\section{Methodology}

\label{sec:method}

\subsection{Overview of Reformulated MARL Framework}

\begin{figure}
  \centering
  \includegraphics[width=0.55\textwidth]{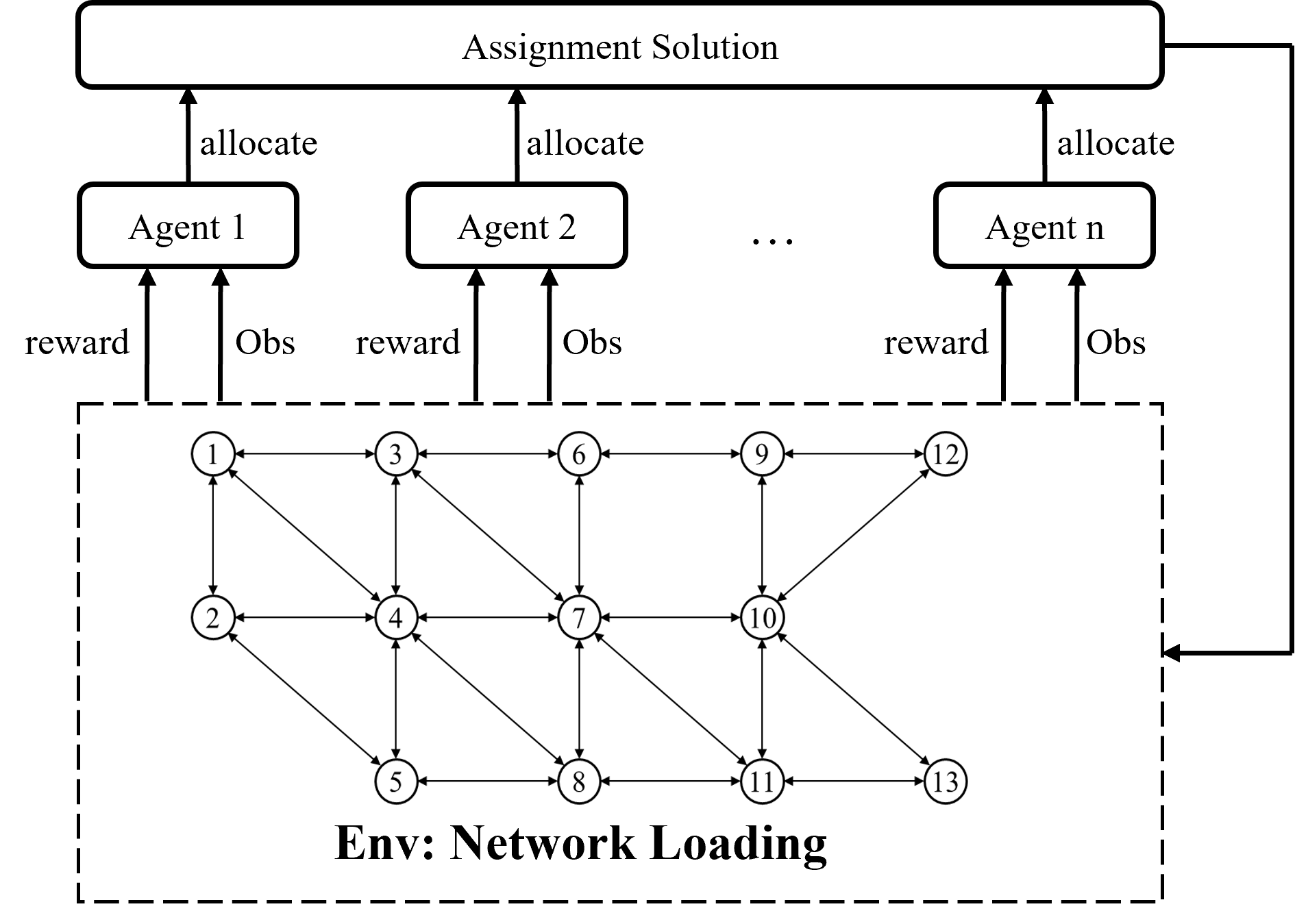}
  \caption{DEC-POMDP framework for traffic assignment}
  \label{fig:MARL_framework}
\end{figure}

Figure~\ref{fig:MARL_framework} illustrates the proposed MARL framework for the traffic assignment problem. Each agent is defined as an OD pair router, with \( n \) corresponding to the number of OD pairs in the network. Agents allocate their OD demand across available routes to optimize individual objectives by maximizing accumulated rewards based on local observations. At each time step \( t \), each agent \( i \) receives local observations \( O_i^t \), which include both intersection-level and route-level information. These observations are inspired by the route choice model and conventional traffic assignment methods and are categorized into static and dynamic features:

\begin{itemize}
    \item \textbf{Static features.}  
    These features remain constant across all time steps and include the origin and destination coordinates, default OD-pair demand, route identifiers, free-flow travel time of available routes, number of links and the average link degree of available routes.
    \item \textbf{Dynamic features.}  
    These features vary with each time step \( t \). They include marginal travel time, recent changes in marginal travel time, travel time observed at the last step, the average volume-to-capacity ratio of links, the number of congested links, and the proportion of demand assigned to each route at the previous step. (All dynamic features are defined at the route level.)
\end{itemize}

Based on local observation \( O_i^t \), the joint action over all agents is denoted by \( \mathcal{A}_t = (A_1^t, A_2^t, \dots, A_n^t) \), where \( A_i^t \) represents the action taken by agent \( i \) at time \( t \). \( \mathcal{A}_t \) serves as an assignment solution, which is then used to perform network loading (i.e., calculating travel costs). Subsequently, the environment provides feedback to each agent \( i \) in the form of a reward \( R_i^t \). The environment then updates each agent's local observation to \( O_i^{t+1} \) for the next step. The objective of each agent is to maximize its discounted accumulated reward, defined as:

\begin{equation}
    J_i(\theta_i) 
= E_{\mathcal{A}_t,\, S_t}
  \biggl[
    \sum_{t=0}^{\infty}
    \gamma^t\, R_i^t\bigl(S_t, \mathcal{A}_t\bigr)
  \biggr],
\end{equation}
where \( \theta_i \) parameterizes the policy \( \pi_{\theta_i}(A_i^t \mid O_i^t) \) for agent \( i \). The policy represents the routing strategy for travelers within the corresponding OD pair, determining the route choice \( A_i^t \) based on the local observation \( O_i^t \).

\subsection{Action Space and Reward}

\subsubsection{Action}  
To cope with issue i in Section \ref{sec: problem statement and analysis}, the action \( A_i \) for each agent \( i \) is defined as the set of proportions assigned to each available route, denoted as \( A_i = \{a_1, a_2, \dots, a_k\} \), where \( k \) is the number of feasible routes for the OD pair. For example, if agent \( i \) corresponds to the OD pair \( 3 \rightarrow 11 \) with two feasible routes, its action \( A_i \) can be \( \{0.2, 0.8\} \), denoting the proportion of OD demand allocate to each route.

The new formulation of action space should satisfy the following probability simplex:

\begin{equation}
    \sum_{i=1}^k a_i = 1, \quad a_i \geq 0, \quad \forall i = 1, 2, ..., k.
\end{equation}

\begin{figure}
  \centering
  \includegraphics[width=0.6\textwidth]{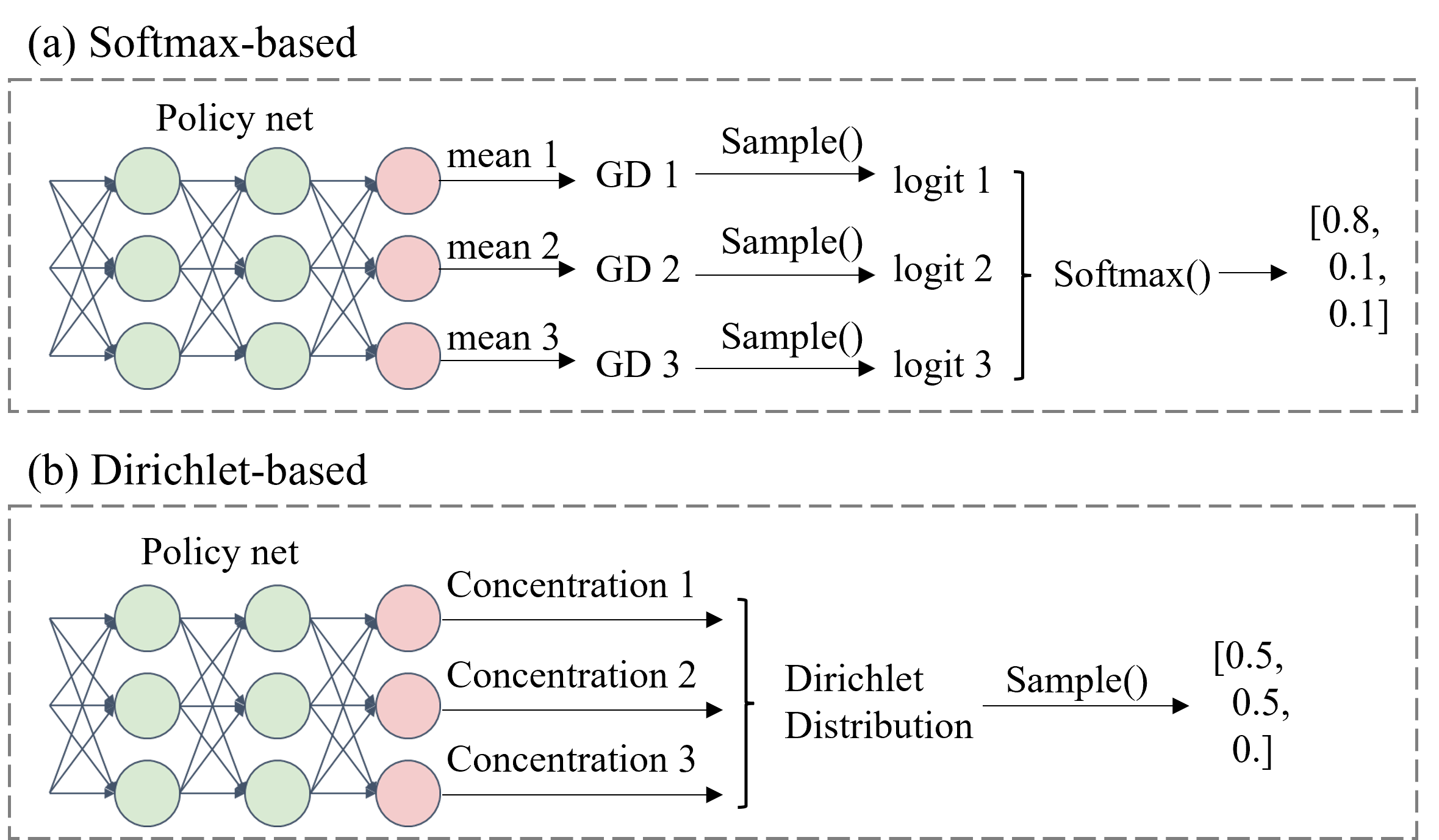}
  \caption{Softmax-based and Dirichlet-based strategy (GD: Gaussian Distribution)}
  \label{fig:action_out}
\end{figure}

\noindent To achieve this goal, softmax-based strategy is an alternative. As shown in Figure~\ref{fig:action_out}(a). It treats the raw outputs of the policy network as logits \( z = [z_1, z_2, ..., z_k] \), and maps them as:

\begin{equation}
    a_i = \frac{\exp(z_i)}{\sum_{j=1}^k \exp(z_j)}, \quad i = 1, 2, ..., k.
\end{equation}

\noindent This formulation guarantees \( \sum_{i=1}^k a_i = 1 \), as the softmax function normalizes the exponentials, and ensures \( a_i > 0 \), since the exponential function is strictly positive.
Despite its simplicity, the softmax transformation cannot represent sparse actions (e.g., \( a_i = 0 \)) due to the strictly positive nature of its output. This limitation reduces the flexibility of the strategy.


Dirichlet-based strategy has the potential to address it, as shown in Figure~\ref{fig:action_out}(b), the  network of Dirichlet-based strategy directly outputs the concentration parameters \( c = [c_1, c_2, ..., c_k] \) using a softplus activation function in its final layer to ensure positivity, then actions are sampled from a Dirichlet distribution with the following probability density function:
\begin{equation}
    p(a \mid c) = \frac{\Gamma\left(\sum_{i=1}^k c_i\right)}{\prod_{i=1}^k \Gamma(c_i)} \prod_{i=1}^k a_i^{c_i - 1},
\end{equation}
where \( a = [a_1, a_2, ..., a_k] \) is the action vector satisfying \( \sum_{i=1}^k a_i = 1 \) and \( a_i \geq 0 \), and \( \Gamma(\cdot) \) is the gamma function.

This Dirichlet-based stragety inherently satisfies the probability simplex constraints while introducing both stochasticity and sparsity in the action vector \( a \) (e.g., allowing some \( a_i \) to be zero).

\subsubsection{Reward}  
\label{sec:reward}

To cope with issue ii in Section \ref{sec: problem statement and analysis}, the reward is derived from the concept of the relative gap, a widely used metric for evaluating convergence in traffic assignment~\cite{boyles2020transportation}. The relative gap quantifies the deviation of the current solution from an optimal state, reaching zero when the flow conditions satisfy either SO or UE. In the UE, it is specified as:
\begin{equation}
    \alpha = \frac{\sum_{e \in E} c_e(x_e) \cdot x_e}{\sum_{(r, s) \in D} k^{rs} d_{rs}} - 1,
    \label{eq:global_RG}
\end{equation}
where \( \alpha \) is the relative gap, \( E \) is the road segments (edges), \( x_e \) is the flow on road segment  \( e \), \( d_{rs} \) is the OD demand between the OD pair \( (r, s) \), \( c_e(x_e) \) is the performance function which maps the flow \( x_e \) on the segment to its corresponding travel time, and \( k^{rs} \) is the travel cost on the shortest path for OD pair \( (r, s) \).

This metric, referred to as the global relative gap, evaluates the assignment across the entire network. Unless otherwise specified, the term relative gap in this paper refers to the global relative gap. For individual agent \( i \), the metric is adapted to a local relative gap based on the flow and demand associated with the agent's specific OD pair \( (r_i, s_i) \):
\begin{equation}
    \alpha_i = \frac{\sum_{p \in P_{r_i s_i}} c_p \cdot f_p}{k^{r_i s_i} d_{r_i s_i}} - 1,
\end{equation}
where \( \alpha_i \) is the local relative gap for agent \( i \), \( P_{r_i s_i} \) is the set of all available paths for OD pair \( (r_i, s_i) \), \( c_p \) is the travel cost for path \( p \), \( f_p \) is the flow assigned to path \( p \), \( k^{r_i s_i} \) is the travel cost on the shortest path, and \( d_{r_i s_i} \) is the demand for \( (r_i, s_i) \).

The reward for each agent is based on the change in its local relative gap between consecutive time steps. At step \( t \), the reward for agent \( i \) is defined as:
\begin{equation}
    R_i^t = 
    \begin{cases} 
    -\alpha_i^1 & \text{if } t = 1 \\
    \alpha_i^{t-1} - \alpha_i^t & \text{if } t > 1
    \end{cases},
\end{equation}
where \( R_i^t \) is the reward for agent \( i \) at time \( t \), and \( \alpha_i^t \) is the local relative gap for agent \( i \) at time \( t \).

\subsection{Action Pruning}

Challenges arise when employing the Dirichlet-based strategy in policy-based reinforcement learning methods. The policy is optimized by computing the gradient of the expected return \( J(\theta) \), defined as:
\begin{equation}
    \nabla_\theta J(\theta) = E_{\pi_\theta} \left[ \nabla_\theta \log \pi_\theta(a \mid s) \cdot \hat{A}(s, a) \right],
\end{equation}
where \( \log \pi_\theta(a \mid s) \) represents the log-probability of the policy, \( s \) denotes the state, and \( \hat{A}(s, a) \) is the advantage function. For the Dirichlet-based strategy, the log-probability of a sampled action \( a \) is given by:
\begin{equation}
    \small
    \log p(a \mid c) = \sum_{i=1}^k (c_i - 1) \log a_i + \log \Gamma\left(\sum_{i=1}^k c_i\right) - \sum_{i=1}^k \log \Gamma(c_i),
\end{equation}
where \( c = [c_1, c_2, ..., c_k] \) denotes the concentration parameters of the Dirichlet distribution.

If any sub-action \( a_i = 0 \), the term \( \log a_i \to -\infty \), rendering the gradient computation undefined and disrupting the policy optimization process. Consequently, even the Dirichlet-based strategy cannot truly achieve sub-actions with exact zero values. Instead, small sub-actions below a predefined threshold (e.g., \( 10^{-7} \)) are clipped to the threshold to prevent this issue.

In traffic assignment, achieving better performance often requires certain sub-actions to be exactly zero. However, this constraint can lead the policy network to excessively reduce logits or concentration values for these sub-actions, potentially resulting in overfitting—a problem similar to the challenges observed in ``Label Smoothing"~\citep{muller2019does,zhang2019bag,szegedy2016rethinking}.

To address this issue, we propose \textbf{action pruning}, inspired by the principles of ``Label Smoothing". This technique modifies the action vector in two steps. First, any sub-actions smaller than a predefined threshold \( \tau \) (e.g., \( 10^{-4} \)) are explicitly set to zero:
\begin{equation}
    a_i'' = 
    \begin{cases}
      0, & \text{if } a_i < \tau, \\
      a_i, & \text{otherwise}
    \end{cases}
\end{equation}
Next, the remaining sub-actions are normalized to ensure they satisfy the probability simplex constraint:
\begin{equation}
    a_i' = \frac{a_i''}{\sum_{j=1}^K a_j''}
\end{equation}
This approach enforces sparsity by setting small sub-actions to zero while ensuring the action vector remains normalized. As the adjustment is applied within the environment, it does not disrupt the agent's policy training or learning process.

\section{Experiment}
\label{sec:experiment}

\subsection{Experiment Settings}
\subsubsection{Urban Transportation Network}

\begin{figure}
  \centering
  \includegraphics[width=0.8\textwidth]{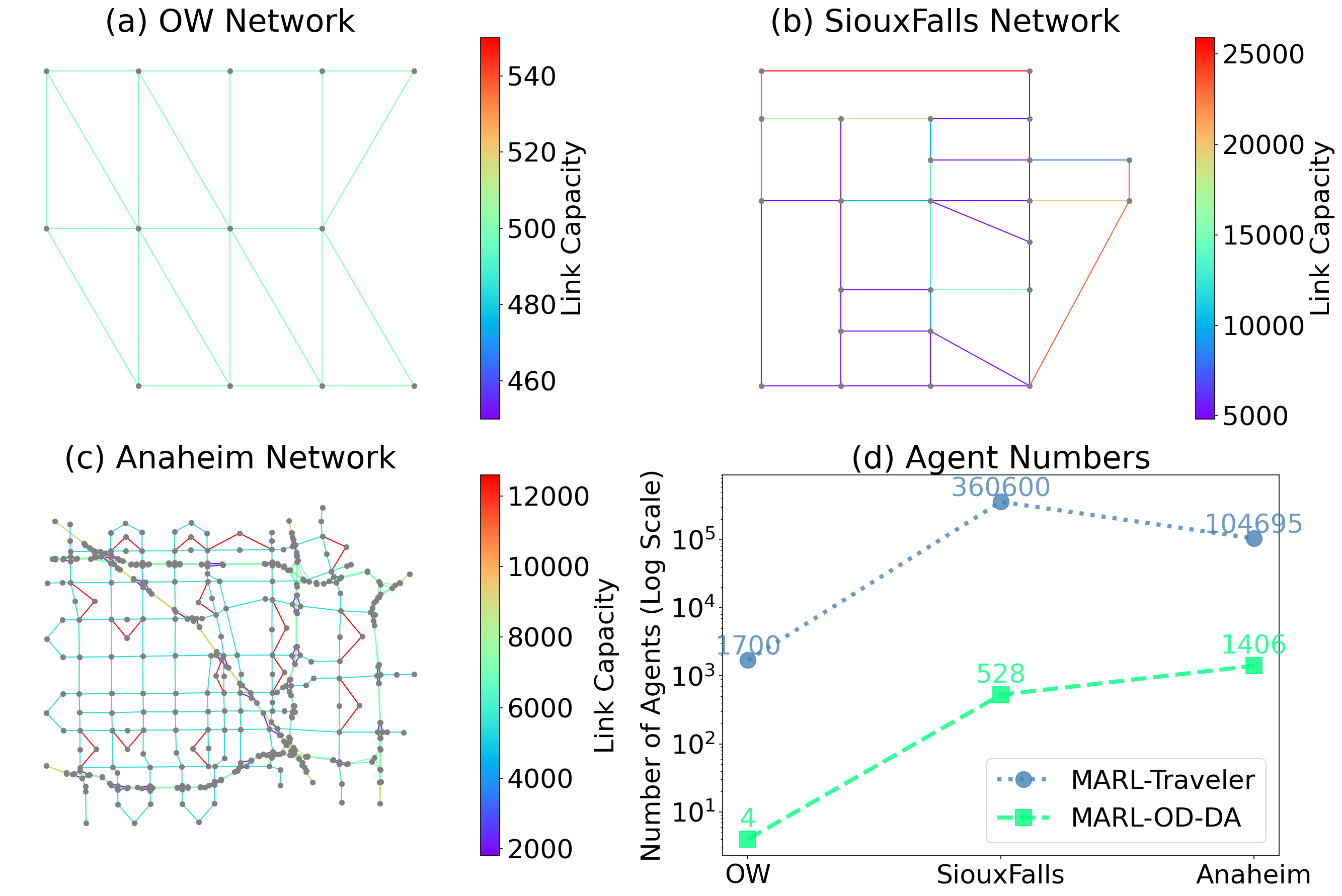}
  \caption{Urban transportation networks and corresponding agent numbers}
  \label{fig:urban_networks}
\end{figure}

The proposed method is evaluated on three widely used urban transportation networks: the Ortuzar and Willumsen (OW) network, the SiouxFalls network, and the Anaheim network~\citep{transportationNetworks,liu2024end,liu2024heterogeneous,de2024modelling}. Their topologies, link capacities, and corresponding agent numbers are shown in Figure~\ref{fig:urban_networks}, with detailed statistics provided in Table~\ref{tab:network_stats}. For all networks, each agent's available route set is derived using the \( k \)-shortest path algorithm~\citep{yen1971finding} or traditional traffic assignment methods, with a maximum of six routes~\citep{boyles2020transportation}.

The OW network, classified as small-sized, is used to compare the proposed algorithm with an existing MARL-based method. In contrast, SiouxFalls and Anaheim are medium-sized networks, with SiouxFalls representing a city-level primary road network due to its high total OD demand (360,600 travelers), and Anaheim reflecting more complex urban systems with its larger number of nodes and links.

\begin{table}
    \centering
    \caption{Detailed information of the urban transportation networks}
    \begin{tabular}{cccccc}
        \toprule
        Network      & Zones & Links & Nodes & OD-pairs & OD-demand \\
        \midrule
        OW           & 13    & 48    & 13    & 4        & 1,700     \\
        SiouxFalls   & 24    & 76    & 24    & 528      & \textbf{360,600}   \\
        Anaheim      & \textbf{38}    & \textbf{914}   & \textbf{416}   & \textbf{1,406}    & 104,694.4 \\
        \bottomrule
    \end{tabular}%
    
    \label{tab:network_stats}
\end{table}

A widely adopted non-linear performance function, the Bureau of Public Roads (BPR) function, is used to calculate the travel time on a given road segment~\citep{boyles2020transportation}:
\begin{equation}
    c_e(x_e) = t_0^e \left(1 + 0.15 \left(\frac{x_e}{w_e}\right)^4\right),
\end{equation}
where \( t_0^e \) denotes the free-flow travel time on \( e \) and \( w_e \) represents the capacity of road segment \( e \). 


To validate the model's stability, variable OD demand is generated by scaling the default OD demand for each OD pair using a random scaling factor, following approaches commonly adopted in previous studies~\citep{liu2024end,liu2024heterogeneous,rahman2023data}. The scaled demand is expressed as:
\begin{equation}
    d_{rs} = \beta_{rs} \cdot d_{rs}^{\text{default}}
    \label {equation:variable_demand}
\end{equation}
where \( d_{rs}^{\text{default}} \) is the default OD demand between origin \( r \) and destination \( s \), and \( \beta_{rs} \) is a uniformly distributed random scaling factor. For the OW and SiouxFalls networks, \( \beta_{rs} \sim U(0.5, 1) \), as their default OD demands represent peak flow under congested conditions. In contrast, the Anaheim network uses \( \beta_{rs} \sim U(0.5, 1.5) \), as its default OD demand reflects less congested conditions, allowing for better representation of traffic fluctuations between moderate and peak volumes, aligning with more realistic dynamics.

\subsubsection{Hyperparameter Settings}

The hyperparameters for the the MARL methods were optimized using grid search. The core settings are summarized in Table~\ref{tab:mappo_hyperparameters}:

\begin{table}[ht]
    \centering
    \caption{Hyperparameter settings}
    \begin{tabular}{cc}
        \toprule
        \textbf{Hyperparameter} & \textbf{Value} \\ \midrule
        Hidden Size & 512 \\ 
        Number of Layers & 3 \\ 
        Learning Rate & $4 \times 10^{-5}$ \\ 
        IPPO Clip Parameter & 0.2 \\ 
        Mini-Batch Size & 512 \\ 
        Discount Factor ($\gamma$) & 0.75 \\ 
        Steps per Episode & 50 \\ 
        Rollout Workers & 5 \\ 
        \bottomrule
    \end{tabular}%
    
    \label{tab:mappo_hyperparameters}
\end{table}



\subsubsection{Evaluation Metrics}
The primary metric is the relative gap, which quantifies the deviation from the optimal solution—a smaller value indicates better performance. For instance, a relative gap of 0.001 roughly represents a 0.1\% deviation. Additionally, the episode average reward is used to assess training performance.

\subsubsection{Baselines}
\textbf{Conventional Methods.}  
In this study, the proposed MARL framework is compared with both conventional and widely used MARL-based methods. Specifically, the conventional methods encompass:

\begin{itemize}
    \item \textbf{Method of Successive Averages (MSA).}  
    An iterative algorithm that updates link flows using a weighted average of previous iterations and current shortest path flows~\citep{mounce2015convergence}.
    \item \textbf{Frank-Wolfe algorithm (FW).}  
    An enhanced version of MSA that employs an adaptive step size and has been widely adopted in practice~\citep{boyles2020transportation}.
\end{itemize}

\textbf{MARL-Based Methods.} All evaluated MARL-based methods use the Independent Proximal Policy Optimization (IPPO) algorithm~\citep{yu2022surprising} for training agents' policies. The methods under consideration comprise:

\begin{itemize}
    \item \textbf{MARL-Traveler.}  
     An altered adaptation of the frameworks as proposed in~\citep{zhou2020reinforcement,shou2022multi} has been implemented, originating from their state-reward design, given the lack of open-source availability. Travelers are modeled as agents with OD locations as observations, route choices as discrete actions, and negative travel costs as rewards. This implementation is developed for comparison due to the lack of open-source code in this domain. This serves as a representative baseline for comparison.
    \item \textbf{MARL-OD-S.}  
    The proposed MARL formulation using a softmax-based strategy. This variant is used as an ablation baseline to assess the impact of replacing the Dirichlet-based strategy.
    \item \textbf{MARL-OD-SA.}  
    The proposed MARL formulation using a softmax-based strategy with action pruning. This variant is designed to evaluate the effectiveness of action pruning method in softmax-based strategy.
    \item \textbf{MARL-OD-D.}  
    The proposed MARL formulation using a Dirichlet-based strategy. This variant is used to isolate the effect of the Dirichlet-based compared to softmax-based strategy.
    \item \textbf{MARL-OD-DA (Ours).}
    The proposed full model combining Dirichlet-based strategy with action pruning.
\end{itemize}

\subsection{Experiment Results}

\subsubsection{Effectiveness Validation Based on Small-Sized Network (OW Network)}

\begin{figure*}[ht]
  \centering
  \includegraphics[width=1\textwidth]{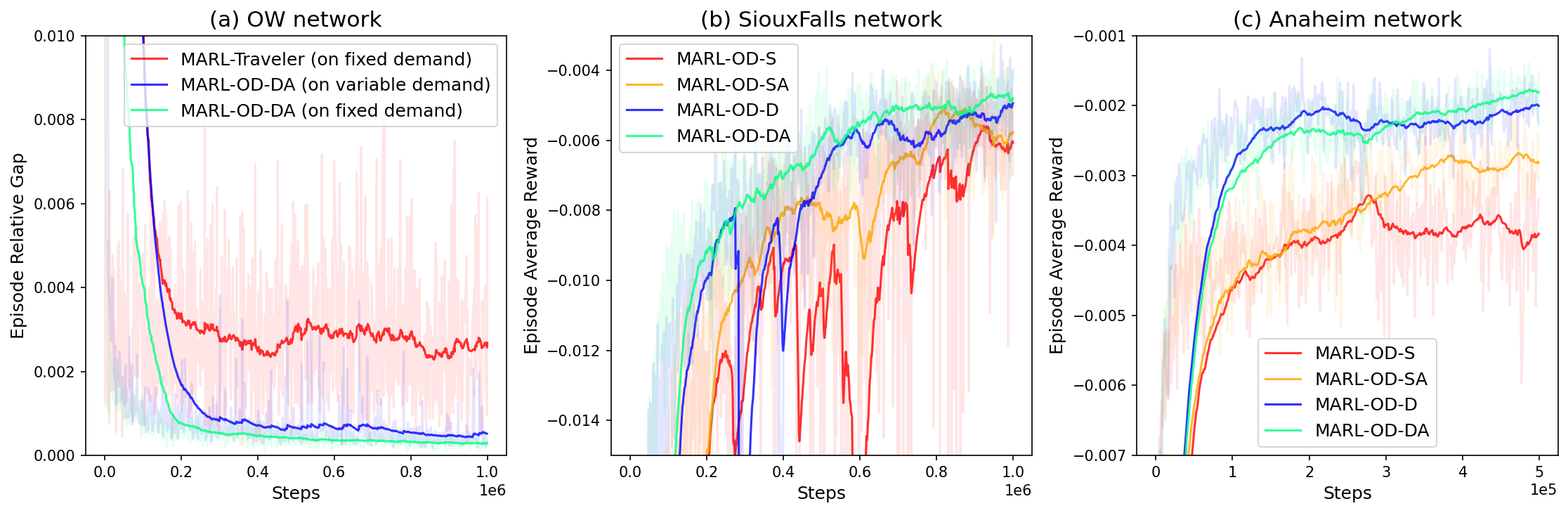}
  \caption{Training curves across three transportation networks}
  \label{fig:train_curve}
\end{figure*}

Figure~\ref{fig:train_curve}(a) illustrates the reduction in the relative gap over training steps for the OW network under two demand scenarios: fixed OD demand, where the demand remains constant as default, and variable OD demand, which evolves as defined in equation~\eqref{equation:variable_demand}. The existing method, \textbf{MARL-Traveler}, is limited to learning under fixed demand scenarios, whereas the proposed method demonstrates significant advantages. Specifically, the proposed method, \textbf{MARL-OD-DA}, achieves a significantly smaller relative gap in both scenarios. Under fixed demand conditions, it not only converges to the lowest relative gap but also maintains strong training stability. Notably, under variable demand conditions, the relative gap achieved by \textbf{MARL-OD-DA} even surpasses the best performance of \textbf{MARL-Traveler} under fixed demand. Note that this comparison is conducted under the UE objective. Since the two methods adopt different reward designs, the episode relative gap to UE is used as a consistent performance metric.

\subsubsection{Scalability Validation Based on Medium-Sized Networks (SiouxFalls and Anaheim Networks)}

Figure~\ref{fig:urban_networks}(d) compares the agent numbers for \textbf{MARL-Traveler} and \textbf{MARL-OD-DA} across the three networks, showing that the proposed framework reduces agent numbers by at least two orders of magnitude. \textbf{MARL-Traveler} is impractical for SiouxFalls and Anaheim networks due to its excessive memory requirements, stemming from the large number of agent networks and replay buffers, which exceed the GPU memory capacity on the experimental hardware (NVIDIA GeForce RTX 3090 Ti). In contrast, the proposed MARL framework demonstrates better scalability and can be applied to these networks.


Figure~\ref{fig:train_curve}(b) and (c) compare the variants of the proposed MARL method (Dirichlet-based vs. softmax-based strategies) on medium-sized networks under a consistent reward structure, highlight that the proposed Dirichlet-based strategy (\textbf{MARL-OD-D}) outperforms the softmax-based strategy (\textbf{MARL-OD-S}). Table~\ref{tab:performance_comparison} further validates the effectiveness of the trained agents in addressing variable OD demand scenarios. The proposed method, \textbf{MARL-OD-DA}, achieves minimum relative gaps of 0.001702 and 0.000869 on the SiouxFalls and Anaheim networks after 50 steps (averaged over five experiments), corresponding to deviations of 0.1702\% and 0.0869\% from the optimal solution. These results demonstrate the framework's scalability to medium-sized networks, its stability in handling variable OD demand, and its reliability in producing high-quality solutions. Note that the experiments on SiouxFalls and Anaheim are conducted under the SO objective, confirming the flexibility of the proposed framework beyond the UE setting.

\begin{table}
    \centering
    \caption{Minimum relative gap in 50 steps (OW-fixed: fixed OD demand; others: variable OD demand)}
    \begin{tabular}{ccccc}
        \toprule
        Method         & OW-fixed & OW & SiouxFalls & Anaheim \\
        \midrule
        MARL-Traveler    & 0.001587   & /                & /                        & /                     \\
        MARL-OD-S     & 0.000418          & 0.001082                & 0.005758                 & 0.001864       \\
        MARL-OD-SA  & 0.000370          & 0.000839                & 0.003266                 & 0.001586       \\
        MARL-OD-D   & 0.000356   & 0.000651         & 0.003903                 & 0.001067              \\
        MARL-OD-DA           & \textbf{0.000218} & \textbf{0.000407} & \textbf{0.001702} & \textbf{0.000869} \\
        \bottomrule
    \end{tabular}%
    
    \label{tab:performance_comparison}
\end{table}

\subsubsection{Reliability Validation of Action Pruning}

Figure~\ref{fig:train_curve}(b) and (c) also demonstrate that both Dirichlet-based and softmax-based strategies benefit significantly from the incorporation of action pruning. With the pruning threshold \( \tau \) set to 0.0001, these variants generally exhibit improved training stability and faster convergence compared to their non-pruned counterparts, with the exception of one case on the Anaheim network where the performance remains comparable.

Table~\ref{tab:performance_comparison} further validates the advantages of action pruning during deployment. For example, \textbf{MARL-OD-DA} achieves a relative gap of 0.001702 on the SiouxFalls network, significantly lower than the 0.003903 reported by \textbf{MARL-OD-D}, corresponding to a reduction of approximately 55.9\%. These results highlight the role of action pruning in improving model reliability and convergence quality.

\subsubsection{Reliability Validation via Comparison with Conventional Methods}

\begin{figure*}[ht]
  \centering
  \includegraphics[width=1\textwidth]{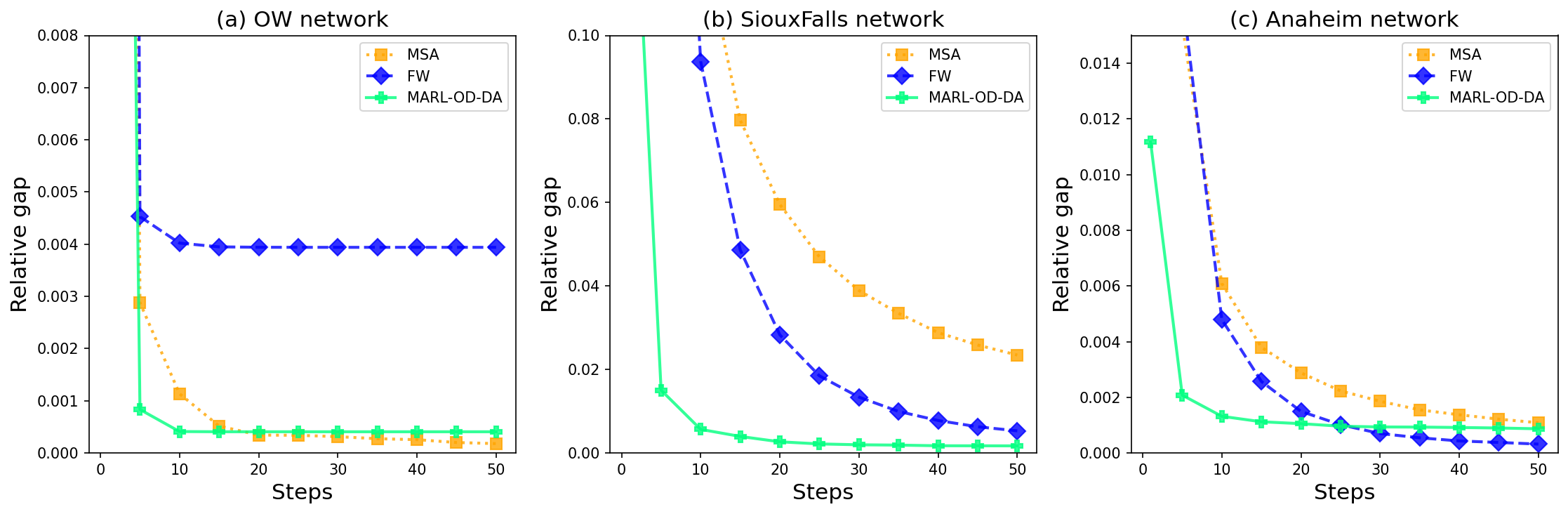}
  \caption{Performance comparison of trained agents and conventional methods across three transportation networks}
  \label{fig:compare_with_conventional}
\end{figure*}

Figure~\ref{fig:compare_with_conventional} compares the proposed MARL method, \textbf{MARL-OD-DA}, with conventional traffic assignment methods, which assume global state knowledge, in terms of solution quality across steps during deployment. For MARL, a step consists of two sub-steps: (1) agents make decisions based on their local observations at step \( t \), and (2) the environment updates by simulating or calculating travel times to produce the state at step \( t+1 \). In contrast, a step in conventional methods involves three sub-steps: (1) computing shortest paths for all OD pairs, (2) optimizing the current assignment along those paths, and (3) simulating or calculating travel times, which aligns with MARL's second sub-step. Among these, shortest path computation in conventional methods becomes increasingly computationally intensive as network size grows, while optimization is relatively fast. Both methods share the same final sub-step of simulating or calculating travel times, which can be highly time-consuming in large networks or when complex traffic flow models or simulation tools are used. This makes the step budget critical for near-real-time applications, where fewer steps are preferred to achieve high-quality solutions.

Figure~\ref{fig:compare_with_conventional} shows \textbf{MARL-OD-DA} achieves higher-quality solutions in fewer steps compared to conventional methods across all three networks. For example, on the SiouxFalls network, \textbf{MARL-OD-DA} achieves a relative gap of approximately \( 0.004626 \) after 10 steps, while the best-performing conventional method, FW, obtains a relative gap of \( 0.092301 \), representing an improvement of approximately \( 94.99\% \). This demonstrates \textbf{MARL-OD-DA}'s efficiency in providing superior solutions in fewer steps, a key advantage for real-time applications. Moreover, unlike conventional methods, MARL does not require global state knowledge, making it more suitable for real-world deployment.

While our method achieves higher-quality solutions with fewer steps, conventional approaches such as \textbf{FW} can eventually catch up given more steps. This is because conventional methods are deterministic numerical optimization algorithms, where each step guarantees a decrease in the objective value. In contrast, MARL relies on neural networks, and fitting errors can cause performance to plateau once a certain accuracy threshold is reached.

\section{Conclusion}
\label{sec:conclusion}


This study proposes a novel MARL paradigm for the traffic assignment problem, redefining agents as OD-pair routers, which reduces the number of agents by two orders of magnitude and significantly enhances scalability. A Dirichlet-based action space is introduced with an action pruning mechanism to enhance the stability of the training and the reliability of convergence. Additionally, the reward function is reformulated based on the relative gap to better guide learning toward optimal solutions. Experimental results on three transportation networks, ranging from small- to medium-scale with variable OD demand, demonstrate the framework's superior scalability and reliability compared to existing MARL-based methods. Furthermore, the proposed approach achieves near-optimal solutions within limited step budgets, making it suitable for near-real-time applications. Ultimately, the proposed approach has the potential to support practical applications such as adaptive routing services, real-time demand management, and policy evaluation in intelligent transportation systems.

A key limitation of this work lies in the use of a simplified analytical traffic assignment framework, which allows for precise theoretical validation but does not fully capture the complexity of real-world traffic dynamics. Future research will extend the framework to high-fidelity simulation environments and large-scale urban networks.

\section*{Acknowledgments}
The work was supported by start-up funds with No. MSRI8001004 and No. MSRI9002005 at Monash University and TRENoP center at KTH Royal Institute of Technology, Sweden.

\section*{AUTHOR CONTRIBUTIONS}
The authors confirm contribution to the paper as follows: study conception and design: Z Ma, L Wang, C Lyu, P Duan, Z Wang, N Zheng; methodology: P Duan, L Wang, Z Ma, Z He, C Lyu; data collection: L Wang, C Lyu, Z Wang, Z He; analysis and interpretation of results: L Wang, C Lyu, Z Ma, Z Wang, P Duan; draft manuscript preparation: L Wang, P Duan, Z Ma. manuscript revision: L Wang, P Duan, Z Ma, C Lyu, N Zheng. All authors reviewed the results and approved the final version of the manuscript.



\bibliographystyle{elsarticle-harv} 
\bibliography{cas-refs}





\end{document}